\documentclass[twocolumn,9pt]{article}

\usepackage{url}
\usepackage{wrapfig}
\usepackage{booktabs}
\usepackage{geometry}
\usepackage{multirow}
\usepackage{amssymb}
\usepackage{amsmath}
\usepackage[utf8]{inputenc}
\newcommand{\footremember}[2]{%
    \footnote{#2}
    \newcounter{#1}
    \setcounter{#1}{\value{footnote}}%
}
\newcommand{\footrecall}[1]{%
    \footnotemark[\value{#1}]%
} 

\title{Google Landmark Retrieval 2021 Competition Third Place Solution}
\author{%
  Qishen Ha \footremember{equal}{Equal contribution. Alphabetical order.} \\ \small LINE Corp%
  \and 
  Bo Liu\footrecall{equal} \footnote{Corresponding author. Email: boli@nvidia.com} \\ \small NVIDIA%
  \and 
  Hongwei Zhang\footrecall{equal} \\ \small LINE Corp%
  }
  
\advance\day by -1
\date{\today}

\usepackage{graphicx}

\begin{document}
\maketitle

\begin{abstract}
We present our solutions to the Google Landmark Challenges 2021, for both the retrieval and the recognition tracks. Both solutions are ensembles of transformers and ConvNet models based on Sub-center ArcFace with dynamic margins \cite{ha2020}. Since the two tracks share the same training data, we used the same pipeline and training approach, but with different model selections for the ensemble and different post-processing. The key improvement over last year is newer state-of-the-art vision architectures, especially transformers which significantly outperform ConvNets for the retrieval task. We finished third and fourth places for the retrieval and recognition tracks respectively.
\end{abstract}

\section{Introduction}

Google Landmark Challenges have been running for the fourth consecutive year \cite{kim2021towards}. In the first two iterations, winning solutions used a combination of global and local features, such as 2019's first place winners in Recognition \cite{2019recognition1st} and Retrieval \cite{2019retrieval1st} tracks. But recent fast advances in state-of-the-art ConvNet models, especially EfficientNet \cite{efficientnet}, resulted in better performance of global feature only models. Neither the first place winner in Retrieval \cite{Retrieval1st} nor Recognition \cite{2020recog1st} last year used local features.

Last year, the third place team (first two authors of this paper were part of the team) in Recognition introduced Sub-center ArcFace with dynamic margins \cite{ha2020} to tackle the extreme class imbalance in the GLDv2 dataset \cite{GLDv2}. It was open sourced and used this year by teams including the first place winner in both tracks \cite{2021winner}. We also used this pipeline for both tracks this year, replacing EfficientNet by newer model architectures. The models are trained the same way for retrieval and recognition. They differ in final model selection for the ensemble and the postprocessing procedure.

\section{Model choices and training recipe}

Sub-center ArcFace with dynamic margins \cite{ha2020}, an improvement over \cite{arcface} and \cite{subcenter} was introduced in last year's recognition competition, and was shown to be effective on the noisy GLDv2 dataset. For both tracks this year, we reuse \cite{ha2020}'s model architectures and training recipe including timm library \cite{timm}, the same augmentations and the progressive training procedure.

Several state-of-the-art vision model architectures have been proposed in the past year. We replaced EfficientNet backbone by EfficientNetV2 \cite{enetv2}, NFNet \cite{nfnet}, ViT \cite{vit} and Swin Transformer \cite{swin} at various model sizes. For transformers, we used a fixed 384 image size. For CNN models, we used progressively larger image sizes (256 $\rightarrow$ 512 $\rightarrow$ 640 or 768).

We found that the CNNs and transformers perform equally well on local GAP scores (recognition metric), but transformers have significantly higher local mAP (retrieval metric) even with the smaller 384 image size. We think this is because transformers are patch-based and can capture more local features and their interactions, which are more important for retrieval tasks.

Besides the new models, we also reused some of our last year's EfficientNet and ReXNet \cite{rexnet} models by finetuning.

\section{The Ensemble}

The retrieval track ensemble consists of 7 models: 3 transformers, 2 new CNNs and 2 old CNNs. Their specifications and local scores are in Table \ref{tab:score1}. Note that ViT large and Swin large have slightly worse local recognition scores than NFNet and EffNetV2 but far superior local retrieval scores.

\begin{table*}[ht!]
\begin{center}
\begin{tabular}{lccccc}
\toprule
Model            & Image size & Total epochs & Finetune 2020 & cv GAP & cv mAP@100 \\ 
&&&& (recognition) & (retrieval)\\\hline
Swin base        & 384        & 60           &                                   & 0.7049               & 0.4442                 \\
Swin large       & 384        & 60           &                                   & 0.6775               & \textbf{0.5161}        \\
ViT large        & 384        & 50           &                                   & 0.6589               & \textbf{0.5633}        \\
ECA NFNet L2     & 512        & 30           &                                   & 0.7021               & 0.3565                 \\
EfficientNet v2l & 640        & 40           &                                   & 0.7129               & 0.4158                 \\
EfficientNet B6  & 512        & 10           & \checkmark              &  &     \\
EfficientNet B7  & 672        & 20           & \checkmark              &  &   \\
\bottomrule
\end{tabular}
\caption{\textbf{Retrieval ensemble's model configuration and local validation scores.} Old CNNs' CV scores are not shown because they are leaky (last year's fold splits were different). }\label{tab:score1}
\end{center}
\end{table*}

The recognition ensemble has 4 more CNNs, shown in Table \ref{tab:score2}. Old CNNs' CV scores are not shown because they are leaky (last year's fold splits were different).

For the retrieval ensemble, adding additional CNNs hurt the score. This is because transformers are more important for retrieval, and adding more CNNs would reduce transformers' relative importance.

\begin{table*}[ht]
\begin{center}
\begin{tabular}{lccccc}
\toprule
Model            & Image size & Total epochs & Finetune 2020 & cv GAP & cv mAP@100 \\ 
&&&& (recognition) & (retrieval)\\\hline
Swin base        & 384        & 60           &                                    & 0.7049                & 0.4442                 \\ 
Swin large       & 384        & 60           &                                    & 0.6775                & 0.5161                 \\ 
ViT large        & 384        & 50           &                                    & 0.6589                & 0.5633                 \\ 
ECA NFNet L2     & 640        & 38           &                                    & 0.7053                & 0.3600                 \\
EfficientNet v2m & 640        & 45           &                                    & 0.7136                & 0.3811                 \\ 
EfficientNet v2l & 512        & 30           &                                    & 0.7146                & 0.4117                 \\ 
EfficientNet B4  & 768        & 10           & \checkmark             &   &    \\ 
EfficientNet B5  & 768        & 20           & \checkmark             &   &    \\ 
EfficientNet B6  & 512        & 20           & \checkmark             & &  \\ 
EfficientNet B7  & 672        & 20           & \checkmark             & &  \\ 
ReXNet 2.0       & 768        & 10           & \checkmark             & &  \\ 
\bottomrule
\end{tabular}
\caption{\textbf{Recognition ensemble's model configuration and local validation scores.} Old CNNs' CV scores are not shown because they are leaky (last year's fold splits were different). }\label{tab:score2}
\end{center}
\end{table*}

\section{Postprocessing}
As shown repeatedly in previous years, postprocessing by way of adjusting raw cosine similarity scores can greatly improve both the GAP and mAP@100 metrics. For both tracks, we designed postprocessing procedures inspired by previous top solutions. They are tuned on local validation and translate well to the leaderboard.

\subsection{Retrieval}
For retrieval competition, one needs to predict up to top 100 matching index images for each query image, sorted by decreasing similarity. 2019 Retrieval's winner used a reranking approach \cite{2019retrieval1st} to bump all the ``positive'' index images to before ``negative'' index images. The positive and negative are defined as whether the model predict the index image and query image to be the same class. This is a very effective approach, but with some drawbacks: the ``positive'' and ``negative'' are predicted by the model hence may be incorrect.

We adopt a softer approach: (1) adjust the similarity scores using a continuous function instead of always bumping positives ahead of negatives, (2) instead of defining positives and negatives using top 1 predictions only, we consider top 3 predictions of both the index image and query image and make adjustments based on the confidence level of the predictions. 

Mathematically, the procedure can be represented by these equations:\\
$$
\text{similarity}= \cos(f_q, f_i)\qquad\qquad\qquad\qquad\qquad\quad
$$
$$
\text{similarity}\  +\!\!= \sum_{j=1}^{3}\sum_{k=1}^{3}g(j, k)\cdot p_j\cdot q_k,\quad \text{if}\  c_j=s_k
$$
$$
\text{similarity}\  -\!\!= \sum_{j=1}^{3}\sum_{k=1}^{3}h(j, k)\cdot p_j\cdot q_k, \quad \text{if}\  c_j=s_k
$$

where $f_q$ and $f_i$ are query image and index image's features; $c_j$ is $j$-th predicted class of the query image and $s_k$ is $k$-th predicted class of the index image; $p_j$ and $q_k$ are the corresponding confidence scores; $g(\cdot)$ and $h(\cdot)$ are rank-based boost function and penalty function respectively, tuned on local validation.

\subsection{Recognition}
For recognition, we incorporated several ideas of last year's top 3 solutions
\begin{itemize}
\item Combine cosine similarity and classification probabilities (3rd place solution 2020 \cite{ha2020})
\item Penalize similarity with non-landmark images (1st place solution 2020 \cite{2020recog1st})
\item Set confidence score of a query image to 0 if its top 5 similarities with non-landmark images are greater than 0.5 (2nd place solution 2020 \cite{2020recog2nd})
\item Penalize CosSim(query, index) for the index class’s count in train set (ours, new)
\end{itemize}

\bibliographystyle{plain}
\bibliography{references}
\end{document}